\documentclass[man,floatsintext,american]{apa7}
\usepackage{amsmath}
\usepackage{algorithm} 
\usepackage{algpseudocode} 
\hypersetup{hidelinks}
\usepackage{csquotes}
\usepackage{float}
\usepackage[style=apa,sortcites=true,sorting=nyt,backend=biber]{biblatex}
\DeclareLanguageMapping{american}{american-apa}
\usepackage{graphicx}
\usepackage{mwe}    
\usepackage{array} 
\usepackage{setspace} 
\usepackage{booktabs} 

\usepackage{lscape}

\begin{document}

\title{Documents are People and Words are Items: A Psychometric Approach to Textual Data with Contextual Embeddings}
\shorttitle{~}


\author{Jinsong Chen}
\authorsaffiliations{Faculty of Education, University of Hong Kong, Hong Kong}





\abstract{This research introduces a novel psychometric method for analyzing textual data using large language models. By leveraging contextual embeddings to create contextual scores, we transform textual data into response data suitable for psychometric analysis. Treating documents as individuals and words as items, this approach provides a natural psychometric interpretation under the assumption that certain keywords, whose contextual meanings vary significantly across documents, can effectively differentiate documents within a corpus. The modeling process comprises two stages: obtaining contextual scores and performing psychometric analysis. In the first stage, we utilize natural language processing techniques and encoder-based transformer models to identify common keywords and generate contextual scores. In the second stage, we employ various types of factor analysis—including exploratory and bifactor models—to extract and define latent factors, determine factor correlations, and identify the most significant words associated with each factor. Applied to the Wiki STEM corpus, our experimental results demonstrate the method's potential to uncover latent knowledge dimensions and patterns within textual data. This approach not only enhances the psychometric analysis of textual data but also holds promise for applications in fields rich in textual information, such as education, psychology, and law.}

\keywords{Contextual embedding, Psychometric modeling, Factor analysis, Large language model, Contextual score}

\maketitle

\section{Introduction}

In recent years, pre-trained large language models (LLMs) have emerged as a driving force in the fields of natural language processing (NLP) and textual data analysis (TDA). Transformer-based architectures \parencite{vaswani2017attention}, such as BERT \parencite{devlin2018bert}, GPT-3 \parencite{brown2020language}, and RoBERTa \parencite{liu2019roberta}, have catalyzed this transformation by delivering state-of-the-art performance across a wide array of tasks, which include keyword extraction, named entity recognition (NER), text summarization, and sentiment analysis among others \parencite[e.g.,][]{chowdhary2020natural,sarkar2019text}. By capitalizing on the vast amounts of data available during pretraining, LLMs effectively capture complex linguistic patterns and relationships, enabling them to generate more accurate and contextually relevant predictions when fine-tuned for specific tasks.

 A key factor contributing to the success of these models in various NLP tasks lies in their ability to convert language into contextual embeddings \parencite{devlin2018bert}. These embeddings encapsulate contextual information and semantic relationships between words, providing a rich representation of text that facilitates deeper understanding and improved performance \parencite[e.g.,][]{ethayarajh2019contextual,rogers2021primer}. They allow for improved document classification based on content \parencite{adhikari2019rethinking} and enhance understanding of entity boundaries and relationships, leading to more accurate NER in textual data \parencite{akbik2018contextual}. Furthermore, contextual embeddings enable LLMs to better comprehend the significance of words and phrases in the text, resulting in more precise keyword extraction \parencite{sahrawat2020keyphrase}. With embeddings, LLMs can leverage pre-trained knowledge and fine-tuning capabilities for specific summarization tasks, creating abstractive and extractive summaries that effectively capture the essence of the original text while remaining concise and easily understandable \parencite{lewis2019bart}. LLMs have also demonstrated significant improvements in sentiment analysis tasks by better grasping the context and nuances of words and phrases, accurately identifying emotions and sentiments expressed within the text \parencite{sun2019utilizing}. With contextual embeddings, transformer-based LLMs have significantly advanced the state of the art in NLP and opened up new avenues for research and application development in the domain of TDA \parencite{rothman2021transformers}.

This research explores a psychometric approach to analyzing textual data by leveraging contextual embeddings generated by LLMs. Unlike many existing tasks, this approach indirectly utilizes contextual embeddings to generate contextual scores, which in turn serve as response data for psychometric analysis such as factor analysis (FA) and classical test theory (CTT). While no fine-tuning is required, generating contextual scores poses a considerable computational burden. However, this burden can be alleviated by utilizing common keywords within a corpus.

The proposed approach offers a natural psychometric interpretation, wherein documents are treated as individuals, words as items, and contextual scores akin to item responses. Under a FA framework, factors are considered latent variables driving variations in response data \parencite{gorsuch2014factor}. In other words, factors are posited as the underlying causes for changes in the meanings of words across different contexts (i.e., documents), as evidenced by the embedding-based contextual scores. An underlying assumption is that certain words, whose contextual meanings vary significantly across documents, can be employed to differentiate documents within a corpus. Such words, often keywords in some documents, are analogous to discriminating items, and their contextual meanings can be captured by contextual scores. Similarly, contextual scores can be used for classical reliability and item analyses. Considering both FA and CTT are foundational to a variety of psychometric tasks such as scale development, construct validation, and measurement invariance, the proposed approach unveils a vast, unexplored landscape for future research, rather than merely introducing a new task. 

The modeling process can be divided into two stages, with the first one for obtaining the contextual scores and the second one for psychometric analysis. In the first stage, we will investigate methods to find common keywords that could potentially characterize specific documents in a corpus. Encoder-based transformer models, such as BERT, are recommended for acquiring contextual embeddings, as they are more efficient than decoder-based transformers. Additionally, several technical aspects need to be explored to reliably obtain contextual scores, which include different versions of BERT, contextual rephrasing, document embeddings, and score computations, among others.

The second stage is akin to psychometric modeling with latent variables. We will assess the normality and linearity of the contextual scores, along with their variations. When the assumptions are satisfied, we will work with the covariance structure rather than the raw data, which simplifies the computation significantly. Factor analytic methods will be employed to estimate the model and obtain the number of factors, factor correlations, and factor loadings. When the number of factors is large, second-order factor analysis will be conducted to obtain and define higher-order factors, which generally have smaller number and simpler interpretations. The results can be re-analyzed with exploratory Bifactor analysis, which offers a natural structure to exclude the interference of a large number of minor factors. Top words with significant loadings can be used to define the general factors substantively. We will also demonstrate classical item analysis with the contextual scores.

The proposed approach, along with the aforementioned procedures, will be experimented with the Wiki STEM corpus, which is abundant with STEM-related content and scientific knowledge, providing a solid foundation to extract and define the factors. The experimental results demonstrate the capabilities of the new approach for psychometric analysis in textual data. Currently, the new approach is primarily exploratory due to the absence of confirmatory theory or hypothesis. However, as knowledge and understanding accumulate over time, the approach can transition towards a more confirmatory tendency, leading to improved interpretation with theoretical support.

\section{Analytical Framework}

\subsection{Applying psychometric concepts to textual data}

Applying psychometric concepts to a corpus containing numerous documents, we can consider each document as an individual possessing specific amounts of different factors. The words within the documents function as items to measure the factors of these documents. In this context, factors are assumed to be latent roots that determine the meaning or contextualization of a specific word in a document, similar to how factors influence an individual's response to a particular item. For example, the word '\textit{bank}' will have different meanings or contextualizations in documents related to geography or finance. Just as two individuals measured by different items might have similar factor scores, two documents composed of different words can also possess similar factors. The corpus's vocabulary encompasses all possible words a document can have, analogous to the hypothesized item pool covering all potential domain behaviors one can be measured with.

To implement this idea, it is crucial to represent the contextual meaning of a specific word in a document in a manner akin to individual item responses. One natural approach is to utilize contextual embeddings, a form of word representation that captures a word's context. These embeddings are generated by LLMs such as ELMo, BERT, or GPT, which are trained on extensive text corpora. When producing contextual embeddings for a word, the LLM considers surrounding words or sentences, encapsulating the semantic and syntactic relationships between words within a given context. This enables the model to generate distinct embeddings for the same word based on its context, allowing for a more accurate representation of the word's meaning in various situations. For instance, take the word "\textit{bank}" in two different sentences: "\textit{She deposited money in the bank}" and "\textit{He sat by the river bank}." Utilizing contextual embeddings, the model will produce different embeddings for "\textit{bank}" in these sentences, reflecting the distinct meanings of the word in each context.

Unlike in psychometrics, where an individual can be measured using any item from an item pool, a document contains only a limited number of words. Nevertheless, we can acquire embeddings for words not present in the original document using LLMs and certain techniques. One straightforward method is to create a new document rephrased as something like "\textit{Find the contextual meaning of 'word' given the following context: 'document'}". With such contextual rephrasing, it is technically feasible, albeit computationally demanding, to obtain embeddings for the entire vocabulary based on every document in the corpus, which will be referred to as conditional contextual embedding (CCE). Although CCEs of the same word based on different forms of rephrasing will be different, they are supposed to be highly correlated, which we will experiment with empirical data. Similar to how responses from related items are highly correlated, CCEs of similar words should also exhibit strong correlation, even if they do not appear in the original document.

\subsection{Obtaining contextual scores as response data}

However, it’s difficult to work with CCE directly due to its high dimensionality. Although one can utilize dimension reduction techniques, such as principal component analysis or singular value decomposition, on CCE, it is challenging to reduce hundreds of dimensions to a scalar similar to item response. This research proposes to adopt the dot product between word-level CCE and document-level embeddings. Two typical methods can be adopted to generate document embeddings with the same vector length of the CCE: 1) Mean-pooling, which is to compute the mean of the token-level embeddings for all the tokens in a document; and 2) Using the embedding for the special token [CLS] generated by BERT as the document-level representation. This token is trained to capture the overall context of the input sequence (i.e., document). Note that there are also other methods to obtain document embeddings that have the same dimensions as word embeddings, such as weighted averaging, which can be evaluated in future research. 

The dot product of CCE and document embedding will be referred to as the contextual score and can be seen as a measure of the similarity between the word and the document. It is a single value effectively captures the degree to which the word's meaning aligns with the overall meaning of the document. Specifically, denote $\textbf{H}_i$ as the embedding for document $i$ and $\textbf{W}_{ij}$ as the CCE for word $j$ conditional on document $i$, one can obtain the contextual score for word $j$ in document $i$ as:

\begin{equation}
    Y_{ij} = \textbf{W}_{ij} \odot \textbf{H}_i.
    \label{cs}
\end{equation}

\begin{figure}[H]
    \begin{center}
    \caption{Generation and Analysis of Contextual Score.}
    \label{fig:cs_proc}    
    \includegraphics[width=.5\linewidth]{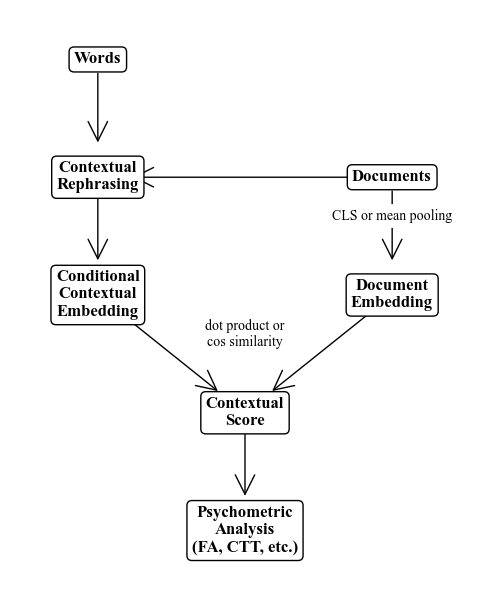}
    \end{center} 
\end{figure}

With the above definition for $\textbf{Y}$, we will obtain contextual scores as response data for psychometric analysis. Note that there are other measures of the similarity between a CCE and a document embedding, such as the cosine similarity, which can also be evaluated as the contextual score in future research. However, the dot product covers a wider range of values, which is more aligned with that of normal distribution. A semantic diagram for the contextual score can be found in Figure \ref{fig:cs_proc}.

\subsection{Factor Analysis based on the Covariance Structure}

\subsubsection{First- and Second- Order Factor Analysis}

In psychometric analysis, FA is a widely used statistical technique to explore the underlying dimensions of psychological constructs, such as intelligence, ability, personality, or attitude. FA allows researchers to examine the relationships between observed variables, also known as items or indicators, and latent constructs or factors. The main idea is to represent the items as a linear combination of factors plus some error terms, which can be expressed in matrix form as:

\begin{equation}
    \textbf{Y} = \textbf{F} \textbf{A}^\text{T} + \textbf{E}.
    \label{fa}
\end{equation}

\noindent where $\textbf{Y}=(Y_{ij})_{N \times J}$, $\textbf{F}=(F_{ik})_{N \times K}$, and $\textbf{E}=(E_{ij})_{N \times J}$ represent the item responses, factors, and error terms, respectively. For textual data, we have $i = 1$ to $N$ representing document, $j = 1$ to $J$ representing word item, and $k = 1$ to $K$ representing factor. The following parameters are of concern: $\textbf{A} =(a_{jk})_{J \times K}$ is the loading matrix, $\textbf{F} \sim N(\textbf{0, C})$ and $\textbf{E} \sim N(\textbf{0, V})$, where $\textbf{C} = (c_{kk'})_{K \times K}$ is the factor covariance matrix and $\textbf{V}= \text{diag}\{v_{jj}\}$ is a $J \times J$ diagonal matrix of the error covariance.

Under the exploratory context, the main purpose of FA is to find a simple and interpretable structure underlying the data, and items that are sufficiently representative of the structure. Moreover, we can work with the covariance structure of the data under the normality assumption, which is usually standardized as a correlation matrix $\textbf{R}$ as:
\begin{equation}
    \textbf{R} = \textbf{ACA}^\text{T} + \textbf{V}.
    \label{sigma}
\end{equation}

\noindent Note that \textbf{C} is also a correlation matrix and there is no need to setup reference loading in this case.

There are two steps in an typical FA. In the first step, one needs to determine the number of factors, which is also referred to as factor extraction. This is done by assuming that the factors are orthogonal to each other with:
\begin{equation}
    \textbf{R} = \textbf{A}_0 \textbf{A}_0^\text{T} + \textbf{V}.
    \label{sigma1}
\end{equation}

\noindent where $\textbf{A}_0$ is the initial loading matrix. There are abundant methods to obtain the number of factors using Eq. \eqref{sigma1}. Established methods include empirical Kaiser criterion (EKC), sequential $\chi^2$ model test, parallel analysis, and Hull’s method \parencite{auerswald2019determine,braeken2017empirical}. But it is unclear if any of them can work well under high-dimensional settings with possibly serious interference in the textual data, which we will examine through empirical data. After factor extraction, factor rotation is conducted to obtain a structure with correlated factors. Rotation aims for a simple and meaningful loading pattern that can be used to define the factors. There are also different methods for factor rotation such as the oblimin and geomin rotations \parencite[e.g.,][]{fabrigar1999evaluating,browne2001overview}.

For a large corpus with a large number of documents and word items, it will likely produce many factors that might be difficult to interpret directly. In this case, we can explore a higher-order (HO) structure with the first-order factors as indicators for second-order factors:
\begin{align}
    \textbf{Y} & = \textbf{F} \textbf{A}^\text{T} + \textbf{E} \\
    \textbf{F} & = \textbf{B} \textbf{Z} + \textbf{D}.
    \label{ho}
\end{align}

\noindent where $\textbf{Z}=(Z_{il})_{N \times L}$ denotes the second-order factors for $l = 1$ to $L$ in document $i$, $\textbf{B}$ denotes the second-order loading matrix, $\textbf{D}=(D_{ik})_{N \times K}$ denotes the second-order error terms, and others are the same as defined in Eq. \eqref{fa}. In addition to parameters in Eq. \eqref{fa}, the following parameters are available: $\textbf{B} = (b_{kl})_{K \times L}$ is the second-order loading matrix, $\textbf{Z} \sim N(\textbf{0, O})$, and $\textbf{D} \sim N(\textbf{0, U})$, where $\textbf{O} = (o_{ll'})_{L \times L}$ is the second-order factor correlation matrix and $\textbf{U} = \text{diag}\{u_{kk}\}$ is a $K \times K$ diagonal matrix.

The covariance structure can be decomposed as:
\begin{align}
    \textbf{R} & = \textbf{ACA}^\text{T} + \textbf{V} \\
    \textbf{C} & = \textbf{BOB}^\text{T} + \textbf{U}.
    \label{ho-sigma}
\end{align}

\noindent For both the first- and second-order levels, similar two-step procedures can be conducted to obtain the number of factors and simple loading patterns for interpretation.

\subsubsection{Bifactor Analysis}

Although the above HO structure provides simpler interpretation with a usually small number of second-order factors, their definitions are coupled with the first-order factors. To achieve independent interpretation across different levels, one can recast the HO structure with Bifactor analysis \parencite[e.g.,][]{jennrich2011exploratory,reise2012rediscovery} by combining the above equations as:
\begin{align}
    \textbf{R} = \textbf{(AB)O(AB)}^\text{T} + \textbf{AUA}^\text{T}+\textbf{V}.
    \label{ho-bif}
\end{align}

\noindent By setting
\begin{align}
   \boldsymbol{\Lambda}_g &= \textbf{AB} \label{SL1} \\
   \boldsymbol{\Lambda}_m &= \textbf{AU}^{\frac{1}{2}},
    \label{SL2}
\end{align}

\noindent one gets a Bifactor structure with multiple general factors that can correlate with each other: 
\begin{align}
    \textbf{R} & = \boldsymbol{\Lambda}_g\textbf{O}\boldsymbol{\Lambda}_g^\text{T} + \boldsymbol{\Lambda}_m\boldsymbol{\Lambda}_m^\text{T}+\textbf{V}.
    \label{bif-sigma}
\end{align}

\noindent where the subscript $g$ and $m$ denotes the elements for the general and minor factors, respectively.

\begin{figure}[H]
    \centering
    \begin{minipage}{0.3\textwidth}
        \centering
        \includegraphics[width=\linewidth]{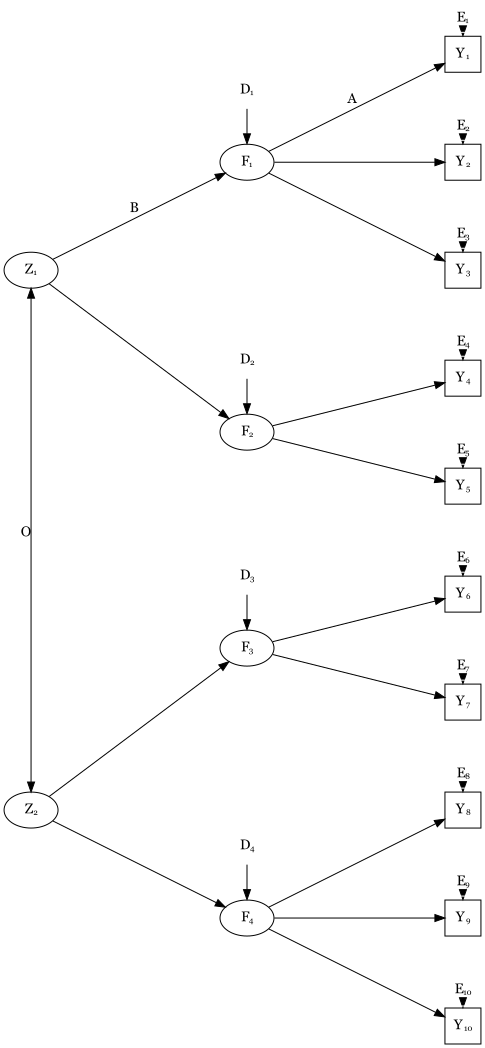}
    \end{minipage}\hspace{1cm}
    \begin{minipage}{0.4\textwidth}
        \centering
        \includegraphics[width=\linewidth]{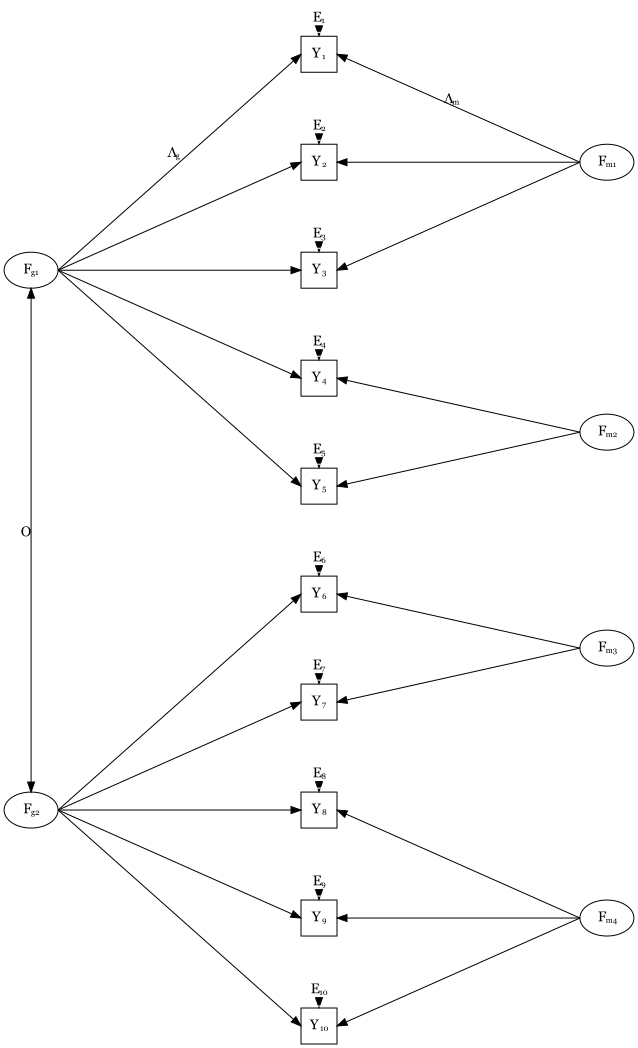} 
    \end{minipage}
    \caption{Higher-order (right) vs. Bifactor (left) Structures.}
    \label{fig:ho-bic} 
\end{figure}

The proportionality constraint in Eqs. \eqref{SL1} and \eqref{SL2} gives the relationship between a HO and constrained Bifactor structures, and is referred to as the Schmid-Leiman (SL) transformation \parencite{schmid1957development}. Although it is often difficult to satisfy the SL transformation perfectly, the proportionality constraint can be sufficiently approximated under many hierarchical exploratory settings. Different from the higher-order solution, the general and minor factors in the Bifactor structure are separated from each other (Figure \ref{fig:ho-bic}). Note that there are the same numbers of general and second-order factors, as well as the same numbers of minor and first-order factors across the two structures. The Bifactor solution offers a viable alternative explanation to textual data with a large number of documents and word items. Due to word similarity, the first-order or minor factors are inevitably subjected to serious interference from word clustering at the local level, making their definitions unreliable. Due to separation however, the general factors under the Bifactor structure can remain relatively stable theoretically, which will be evaluated with the experiment data. 

\subsection{Data Preprocessing before contextual score}

The number of words in a corpus's vocabulary is usually vast. Analyzing contextual scores of the entire vocabulary is computationally intensive and unnecessary, since most words will not be useful for psychometric analysis. Similar to useful items in traditional psychometric models, useful words should characterize documents and discriminate between those with high and low factor scores. This implies that we need to focus only on common keywords shared by at least some documents, significantly reducing the number of words to analyze. 

To obtain common keywords in a corpus, one can first preprocess the text with NLP techniques such as tokenization, removing stop words, and lemmatization \parencite{jurafsky2008speech,feldman2007text}. Lemmatization reduces words to their base or root form, known as the lemma. It analyzes the morphological structure of the word and applying linguistic rules to transform it into its base form by removing any inflectional affixes (such as prefixes, suffixes, or infixes) that indicate tense, gender, case, or other grammatical properties. After preprocessing, one can apply various keyword extraction techniques to identify the most relevant and significant keywords in each document \parencite[e.g.,][]{bird2009natural,sarkar2019text}. Some popular methods that can be readily implemented using the NLTK library \parencite{loper2002nltk} include: (1) Term frequency-inverse document frequency (TF-IDF), which calculates the importance of a term in a document relative to its frequency in a collection of documents (corpus); (2) TextRank, which is a graph-based algorithm ranks the words based on their importance in a graph with words as nodes and edges representing their co-occurrence relationships in the text; (3) Rapid automatic keyword extraction (RAKE), which identifies keywords by computing a score for each word based on its frequency and the degree to which it appears with other words.

After extracting the top $n$ keywords from each document, one can identify the shared keywords across the documents by finding the intersection of the keyword sets. One can also calculate the occurrences of each shared keyword across the documents to determine the most common keywords in the corpus. The second approach is more appropriate to find useful keywords to differentiate between documents. This research will set $n=10$, which provides a more detailed representation of the main topics and is often used in exploratory analyses, summaries, or visualizations.

Two data issues are relevant to psychometric analysis with contextual scores. First, highly correlated word items can result in serious multicollinearity and are often associated with problems such as unstable estimates, difficulty in identifying unique factors, and inflated standard errors \parencite{kline2023principles}. To address these problems, it is often recommended to remove highly correlated items from the analysis. This can be done by examining the correlation matrix of the items and identifying pairs of items with high correlations (e.g., above a predetermined threshold). One item from each pair can be removed based on theoretical considerations, item content, or other criteria. Second, covariance structure analysis assumes that the observed variables have a multivariate normal distribution. Non-normality can lead to biased estimates of factor loadings and inaccurate inferences about the underlying factor structure. However, FA is generally considered to be robust to moderate violations of normality when the sample size is large.

\section{Experiment}

\subsection{Stage One: Obtaining Contextual Scores}

The Wiki STEM Corpus is a collection of articles and information from Wikipedia that is focused specifically on topics related to science, technology, engineering, and mathematics. The dataset is available in Kaggle, and a step-by-step guide on creating this dataset from Wikipedia articles can be found in this link: https://www.kaggle.com/code/nbroad/create-science-wikipedia-dataset. By narrowing the focus of the corpus to STEM topics, the content is tailored to scientific knowledge, providing a solid foundation to extract and define the factors.

Due to the limitation of BERT type models, documents within 500 tokens are selected for analysis. Meanwhile, documents smaller than 50 tokens are removed due to little information they can provide. Moreover, only the first 20,000 documents are used to reduce computational burden, which already gives most keywords one will get using the entire dataset. As compared with TextRank and RAKE, TF-IDF can extract more keywords under the same conditions and is adopted. After preprocess the text with stop word removal and lemmatization, $1,423$ words are obtained by setting both $n$ and occurrences as $10$. The BERT large model is adopted, which, although more demanding computationally, gives substantially different results as compared with the BERT base model.

Contextual scores will be obtained according to Eq. \eqref{cs}. For document embedding, both the mean-pooling and [CLS] methods will be adopted for comparison. To test different forms of contextual rephrasing, 50 keywords and 1000 documents are randomly selected for experiment. As shown in Table \ref{tab:1}, contextual scores obtained using various forms are highly correlated and will give essentially identical results under the covariance structure analysis.

\begin{table}[H]
    \singlespacing
     \centering
     \caption{Average Contextual Score Correlations between Different Rephrasing Forms}
     \label{tab:1}
    \setlength{\tabcolsep}{5pt} 
    \begin{tabular}{p{8cm}cccccc}
    \toprule
    Form & 1 & 2 & 3 & 4 & 5 & 6 \\
     \hline
    1. \textit{"find the contextual meaning of '\{word\}' given the following context: \{document\}"} &  -	& 0.990	& 0.926 & 0.979 &	0.955 & 0.956 \\
    2. \textit{"find the meaning of '\{word\}' given the following context: \{document\}"} &  0.993&	-	&0.915&	0.981&	0.949&	0.957 \\
    3. \textit{"find the embedding of '\{word\}' given the following context: \{document\}"} &  0.967&	0.967&	-&	0.887&	0.972&	0.876 \\
    4. \textit{"what is the meaning of '\{word\}' given the following context: \{document\}"}
   &  0.990&	0.988&	0.953&	-&	0.942&	0.962 \\
    5. \textit{"what is the embedding of '\{word\}' given the following context: \{document\}"}
    &  0.974&	0.972&	0.982&	0.972&	-&	0.921 \\
    6. \textit{"the word is '\{word\}' and the context is: \{document\}"} & 0.977&	0.981&	0.956&	0.971&	0.960&	- \\
    \bottomrule
    \end{tabular}
    {\\ \textit{Note}. Lower triangle based on mean-pooling embedding;  upper triangle based on [CLS] embedding.}
\end{table}

\begin{figure}[H]
     \centering
    \caption{Scatter and Density Plots for Typical Words with Different Levels of Occurrences.}
    \label{fig:1}    
    \includegraphics[width=.8\linewidth]{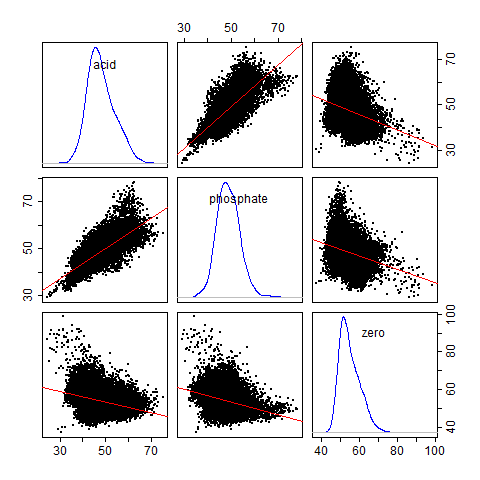} \\
    {\textit{Note}. High levels: \textit{'acid'}, moderate levels: \textit{'phosphate'}, low levels: \textit{'zero'}.}
\end{figure}

Using high performing computing with three Nvidia V100 GPUs, it took about four days to obtain the contextual scores for the 1,423 words across the 20k documents using the BERT large model with Form 1 rephrasing. For the contextual scores of most words, the nonnormality (i.e., skewness and kurtosis) are no more than moderate and the bivariate relationships appear to be linear. For the [CLS]-based scores, however, the average word-pairs correlation is about .63 and most pairs are highly correlated (.8 or above), which will result in serious multicollinearity and are not appropriate for factor analysis. For the mean pooling-based scores, the average correlation is about .17 and most correlations are about moderately or below, which suggests that the method can be used for further analysis. The scatter and density plots for three typical words with high, moderate, and low occurrences can be found in Figure \ref{fig:1}, demonstrating normality and linearity within moderate violation, which is suffice for FA with large sample size \parencite[e.g.,][]{fabrigar1999evaluating,muthen1985comparison}.

\begin{figure}[H]
    \begin{center}
    \caption{Density Plots for Full and Split Data.}
    \label{fig:11}    
    \includegraphics[width=.6\linewidth]{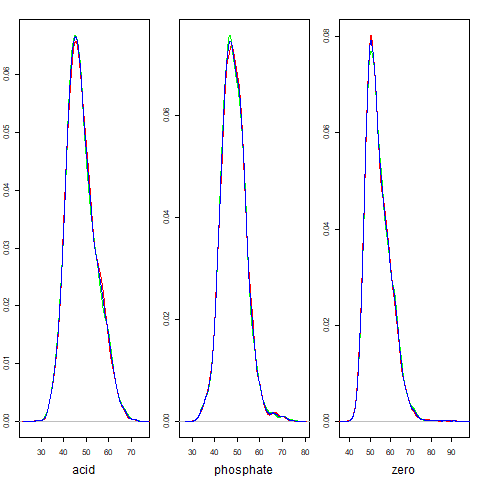}
    \end{center} 
    {\textit{Note}. Blue - full, green - first half, red - second half.}
\end{figure}

To more thoroughly assess the score distributions, the data is randomly divided into two equal parts. As depicted in Figure \ref{fig:11}, the density plots for each half and the complete dataset are essentially overlapped, indicating that a word's characteristics, associated with the distribution, remain stable and unaffected by specific documents. This finding aligns with the notion of independence between items and individuals, a crucial assumption for estimating item and person parameters in psychometric modeling.

\begin{figure}[H]
     \centering
    \caption{Scatter and Density Plots for Highly Correlated Words.}
    \label{fig:2}    
    \includegraphics[width=.8\linewidth]{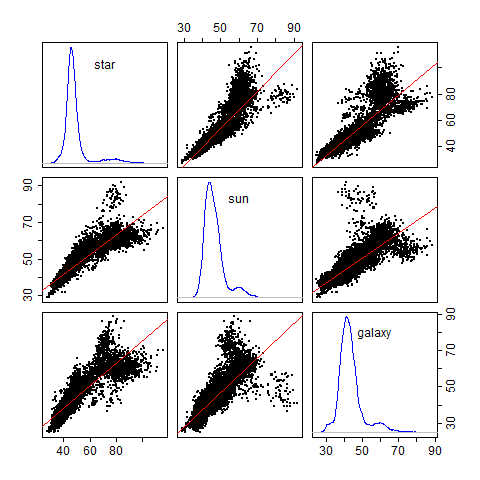} \\
    {\textit{Note}. All correlations are above .8.}
\end{figure}

Many word pairs are highly correlated. The scatter and density plots for three highly correlated words can be found in Figure \ref{fig:2}, demonstrating the close relationships among three words \textit{star}, \textit{sun}, and \textit{galaxy}, which makes sense substantively. In addition to the full data, data filtered by removing the low-occurrence words in highly correlated word pairs ($r>.8$) are also analyzed to evaluate the possible influence of multicollinearity.

\subsection{Stage Two: Conducting Factor Analysis}

\subsubsection{First- and Second-Order Factor Analysis}

\begin{figure}[H]
    \begin{center}
    \caption{Scree Plot of First-order FA for Full Data.}
    \label{fig:3}    
    \includegraphics[width=.8\linewidth]{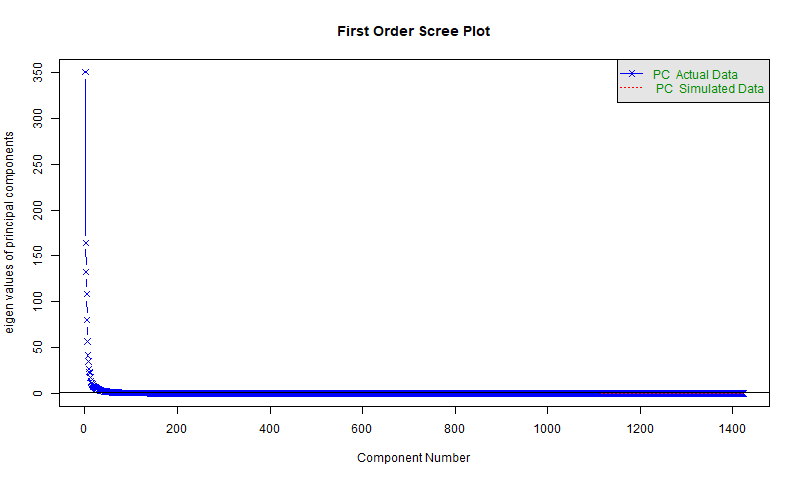}
    \end{center}
\end{figure}

Summary of analysis for the full and filtered data can be found in Table \ref{tab:ho-fa}. In the first step of FA, 60 and 53 factors have been extracted using parallel analysis for the full and filtered data, respectively. Other methods, such as the EKC, sequential $\chi^2$ model test, and Hull’s method, either won't be able produce any result or yield a much larger number that seems unreasonable. The corresponding scree plots for the full and filtered data can be found in Figure \ref{fig:3} and Appendix A, respectively. Both show a clear elbow pattern, although the cutting point is somewhat ambiguous. Meanwhile, these 60 and 53 factors account for over $85\%$ of the total covariance in the data. Given that the typical value in traditional FA settings is about $40\% \sim 70\%$, the number suggests a potential overextraction of factors.

\begin{table}[H]
    \singlespacing
    \begin{center}
     \caption{Summary of Higher-order FA}
     \label{tab:ho-fa}
    \begin{tabular}{lcccc}
    \toprule
& \multicolumn{2}{c}{First-order} &  \multicolumn{2}{c}{Second-order}\\
     \hline
&	Full	&	Filtered	&	Full	&	Filtered	\\    
\# of item	&	1423	&	969	&	60	&	53	\\
Mean item correlation	&	0.174	&	0.184	&	-	&	-	\\
Mean abs. item correlation	&	0.254	&	0.238	&	-	&	-	\\
\# of factors (parallel analysis)	&	60	&	53	&	13	&	13	\\
Mean factor correlation	&	0.062	&	0.064	&	0.051	&	0.046	\\
Mean abs. factor correlation	&	0.104	&	0.101	&	0.076	&	0.076	\\
Total variance ratio	&	0.887	&	0.860	&	0.419	&	0.415	\\
\# of loading per factor (||>.3)	&	12.817	&	11.604	&	4.615	&	4.538	\\
\# of loading per factor (||>.5)	&	2.383	&	1.981	&	1.692	&	1.692	\\
    \bottomrule
    \end{tabular}
    \end{center}
\end{table}

After determining the number of factors, factor rotation is performed to obtain factor correlations and a simplified loading pattern. To conserve space, all correlation and loading estimates for both full and filtered data are presented in the supplemental materials (\href{https://osf.io/sm67u/}{https://osf.io/sm67u/}). The top 20 keywords for each factor, ranked according to their loading magnitude, can be also found in supplemental materials, together with their corresponding loading values. Factor correlations range from near zero to small, indicating the independence of the factors. Considering the average numbers of significant loadings per factor based on the conventional cutoff of 0.3 (Table \ref{tab:ho-fa}), many factors have no more than 10 items (i.e., keywords) for operational definition. Although it is not a problem in traditional FA, the number of keywords might not be sufficiently reliable to define a factor under a textual environment. Note that most loading magnitudes are smaller than .5, further adding to the issue of reliability.

\begin{figure}[H]
    \begin{center}
    \caption{Scree Plot of Second-Order FA for Full data.}
    \label{fig:4}    
    \includegraphics[width=.8\linewidth]{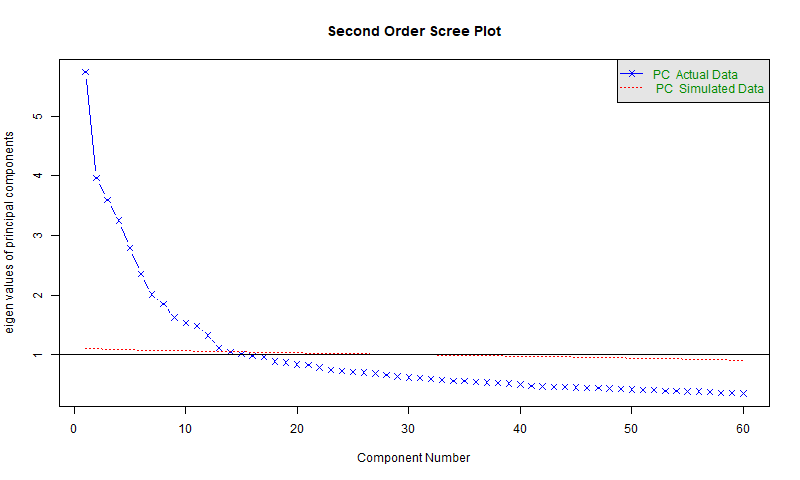}
    \end{center}
\end{figure}

Considering that the number of extracted factors and total covariance accounted for are large, a second-order FA has been conducted with the above factors as indicators and following similar two-step procedure. Note that 13 factors have been extracted using parallel analysis for both the full and filtered data, suggesting the reliability of factor extraction at the second order. The scree plots for the full and filtered data can be found in Figure \ref{fig:4} and Appendix B, respectively. These 13 factors account for about $41\%$ of the total covariance in the first-order factors, which is relatively small but not uncommon in traditional FA settings. Correlations between second-order factors for the full and filtered data can be found in Table \ref{tab:5} and Appendix C, respectively, which shows that the factors are weakly correlated in general. Meanwhile, both second-order loading matrices (supplemental materials) are sparse based on a .3 cutoff, and the average numbers of significant loadings per factor are rather small, suggesting likely unreliable factor definition. Note that the reliability issues of the first- and second-order factors are coupled to add to the complexity.

\begin{table}[H]
    \singlespacing
    \begin{center}
     \caption{Second-order Factor Correlations for Full Data}
     \label{tab:5}
    \setlength{\tabcolsep}{3pt} 
    \small
    \begin{tabular}{lcccccccccccc}
    \toprule
 &	$Z_1$	&	$Z_2$	&	$Z_3$	&	$Z_4$	&	$Z_5$	&	$Z_6$	&	$Z_7$	&	$Z_8$	&	$Z_9$	&	$Z_{10}$	&	$Z_{11}$	&	$Z_{12}$	\\
     \hline
$Z_2$	&	-0.055	&		&		&		&		&		&		&		&		&		&		&		\\
$Z_3$	&	-0.038	&	0.073	&		&		&		&		&		&		&		&		&		&		\\
$Z_4$	&	-0.030	&	0.107	&	0.159	&		&		&		&		&		&		&		&		&		\\
$Z_5$	&	0.015	&	0.064	&	0.055	&	0.021	&		&		&		&		&		&		&		&		\\
$Z_6$	&	0.029	&	-0.002	&	0.118	&	-0.031	&	0.163	&		&		&		&		&		&		&		\\
$Z_7$	&	0.011	&	0.094	&	0.075	&	0.046	&	0.075	&	-0.011	&		&		&		&		&		&		\\
$Z_8$	&	0.010	&	0.033	&	0.087	&	0.087	&	0.126	&	0.236	&	0.039	&		&		&		&		&		\\
$Z_9$	&	0.045	&	0.041	&	0.155	&	0.196	&	-0.065	&	0.023	&	-0.118	&	0.036	&		&		&		&		\\
$Z_{10}$	&	-0.005	&	0.178	&	0.210	&	0.034	&	0.098	&	0.136	&	-0.004	&	0.110	&	0.032	&		&		&		\\
$Z_{11}$	&	0.171	&	-0.034	&	-0.006	&	-0.124	&	0.120	&	0.059	&	0.002	&	0.027	&	0.011	&	0.085	&		&		\\
$Z_{12}$	&	-0.119	&	0.081	&	0.034	&	0.055	&	0.011	&	0.156	&	-0.180	&	0.139	&	0.109	&	0.069	&	0.035	&		\\
$Z_{13}$	&	0.149	&	0.201	&	0.001	&	0.122	&	0.083	&	0.014	&	-0.009	&	-0.014	&	0.151	&	0.005	&	-0.149	&	0.056	\\
    \bottomrule
    \end{tabular}
    \end{center}
\end{table}

\subsubsection{Bifactor Analysis}

Exploratory Bifactor analysis has been adopted to re-interpret the higher-order structure using the SL transformation in Eqs. \eqref{SL1} and \eqref{SL2}. Summary of analysis for the full and filtered data can be found in Table \ref{tab:bifa}. For both cases, the general factors account for most of the explained variance at the first order. The average numbers of loadings per factor are large, with many loading magnitudes of .5 or above, for the general factors. In contrast, the average numbers of loadings per factor are small, with few loading magnitudes of .5 or above, for the minor factors. The results suggest that the general factors can be reliably defined with a large number of significant keywords although the minor factors cannot, which is consistent with the separation of the two types of factors in a Bifactor model.

\begin{table}[H]
    \singlespacing
    \begin{center}
     \caption{Summary of Bifactor Analysis}
     \label{tab:bifa}
    \begin{tabular}{lcccc}
    \toprule
& \multicolumn{2}{c}{General Factor} &  \multicolumn{2}{c}{Minor Factor}\\
     \hline
&	Full	&	Filtered	&	Full	&	Filtered	\\
\# of factors	&	13	&	13	&	60	&	53	\\
Total variance ratio	&	0.617	&	0.572	&	0.269	&	0.287	\\
\# of loading per factor (||>.3)	&	158.615	&	105.231	&	6.083	&	5.453	\\
\# of loading per factor (||>.5)	&	38.308	&	17.077	&	0.617	&	0.679	\\
    \bottomrule
    \end{tabular}
    \end{center}
\end{table}

Top words are the key to define the factors substantively. The top 30 keywords for each general factor are listed in Table \ref{tab:7} and Appendix D, while their corresponding loading values can be found in Table \ref{tab:8} and Appendix E, for the full and filtered cases, respectively. The magnitudes of all above loadings are beyond .3, and most are beyond .5, suggesting the reliability of the measurement. Labels for the general factors suggested by ChatGPT based on top 30 keywords and corresponding loadings can be found in in Table \ref{tab:6}, which seems to align well with STEM-related constructs and represent knowledge covered in the Wiki STEM Corpus. Note that factors with most negative loadings are labeled with \textit{"Inverse"}. It is interesting to see how the labels and wording vary across the two lists, which reflects the changes of the keywords in the two datasets. Finally, it is worthy noting that correlations of the general factors in the Bifactor analysis share the same correlations of the second-order factors in the HO structure.

\begin{landscape}
\begin{table}[H]
     \centering
     \caption{Top 30 Words of General Factors for Full Data}
     \label{tab:7}
    \renewcommand{\arraystretch}{1}
    \setlength{\tabcolsep}{0pt} 
    \footnotesize
    \begin{tabular}{ccccccccccccc}
    \toprule
$F_1$	&	$F_2$	&	$F_3$	&	$F_4$	&	$F_5$	&	$F_6$	&	$F_7$	&	$F_8$	&	$F_9$	&	$F_{10}$	&	$F_{11}$	&	$F_{12}$	&	$F_{13}$	\\
     \hline
sulfur	&	main	&	byte	&	theorem	&	nucleosomes	&	algae	&	rna	&	ligament	&	cytokine	&	force	&	antifungal	&	transfer	&	hiv	\\
zinc	&	constellation	&	software	&	conjecture	&	cholesterol	&	mushroom	&	motif	&	nerve	&	cysteine	&	energy	&	antipsychotic	&	function	&	human	\\
chloride	&	major	&	command	&	inequality	&	orthologs	&	marine	&	transcription	&	spinal	&	cytochrome	&	pressure	&	antibiotic	&	acceptor	&	leukemia	\\
sugar	&	star	&	file	&	abelian	&	lymphoma	&	plant	&	replication	&	dorsal	&	microrna	&	stress	&	antagonist	&	ligases	&	condensation	\\
nitrogen	&	gamma	&	instruction	&	proof	&	nucleotide	&	soil	&	exon	&	artery	&	protein	&	event	&	antioxidant	&	dependent	&	antigen	\\
amino	&	au	&	code	&	algebra	&	streptomyces	&	fungus	&	variant	&	duct	&	partial	&	heat	&	selective	&	transferase	&	insulin	\\
vinyl	&	dwarf	&	program	&	let	&	bacillus	&	food	&	initiation	&	muscle	&	receptor	&	shock	&	antidepressant	&	enzyme	&	rat	\\
nickel	&	giant	&	server	&	infinite	&	nucleolar	&	colony	&	trna	&	cord	&	may	&	dynamic	&	medication	&	ligase	&	patient	\\
glucose	&	companion	&	java	&	math	&	gastrointestinal	&	fish	&	repeat	&	vascular	&	gtpase	&	delay	&	drug	&	local	&	dodecahedron	\\
benzoate	&	kappa	&	processor	&	finite	&	nucleoside	&	organism	&	stranded	&	vein	&	superfamily	&	moment	&	fda	&	action	&	forming	\\
chemistry	&	asteroid	&	library	&	property	&	vaginal	&	fungal	&	genetic	&	canal	&	membered	&	wave	&	therapy	&	initiation	&	atomic	\\
magnesium	&	catalogue	&	package	&	manifold	&	cannabinoid	&	fruit	&	transcript	&	anterior	&	membrane	&	electrical	&	antiviral	&	residue	&	disease	\\
amine	&	ngc	&	chip	&	algebraic	&	dystrophy	&	sea	&	dna	&	neck	&	macrophage	&	magnetic	&	dose	&	functional	&	coordination	\\
alcohol	&	spectral	&	programming	&	lie	&	pyridoxal	&	fungi	&	sequence	&	bone	&	antigen	&	voltage	&	effect	&	pump	&	fetal	\\
bond	&	cluster	&	store	&	matrix	&	pyridine	&	fossil	&	rib	&	tendon	&	iv	&	change	&	safety	&	pathway	&	cancer	\\
carbon	&	milky	&	security	&	theory	&	carbohydrate	&	insect	&	strand	&	cortex	&	additive	&	loss	&	treatment	&	bound	&	hepatitis	\\
ammonia	&	saros	&	computer	&	every	&	hexagonal	&	parasite	&	copy	&	cerebral	&	polymerase	&	velocity	&	trial	&	transport	&	mutant	\\
aminoacyl	&	spectrum	&	linux	&	define	&	mycobacterium	&	animal	&	site	&	motor	&	gene	&	electric	&	agent	&	translation	&	diabetes	\\
benzene	&	jupiter	&	bit	&	law	&	eridanus	&	yeast	&	translation	&	cervical	&	mutant	&	potential	&	antibody	&	binding	&	formation	\\
arsenic	&	hd	&	http	&	continuous	&	lymph	&	bacterial	&	interaction	&	neural	&	histone	&	noise	&	potent	&	exchange	&	composition	\\
phosphorus	&	planet	&	service	&	dimension	&	lysine	&	leaf	&	mrna	&	surgery	&	kinase	&	wind	&	medicine	&	metabolism	&	mutation	\\
oxide	&	component	&	format	&	topology	&	plasmodium	&	ocean	&	genome	&	gland	&	total	&	rate	&	brand	&	transporter	&	inflammation	\\
carbonyl	&	minor	&	cpu	&	polynomial	&	pyruvate	&	bacterium	&	coding	&	fiber	&	biogenesis	&	cause	&	exposure	&	free	&	gene	\\
oxidation	&	super	&	machine	&	dimensional	&	ubiquinone	&	flower	&	structure	&	smooth	&	microtubule	&	temperature	&	clinical	&	name	&	halide	\\
aluminium	&	survey	&	key	&	sum	&	nonsteroidal	&	gram	&	terminal	&	skull	&	alternate	&	current	&	psychedelic	&	interact	&	tumor	\\
acid	&	beta	&	database	&	differential	&	plexus	&	wood	&	entry	&	convex	&	genome	&	transition	&	analgesic	&	product	&	orbital	\\
oxygen	&	massive	&	memory	&	rule	&	keratin	&	culture	&	version	&	medial	&	equal	&	particle	&	inhibitor	&	polygon	&	tert	\\
nitrate	&	magnitude	&	unix	&	hypothesis	&	epinephelus	&	bacteria	&	box	&	lateral	&	biosynthesis	&	speed	&	action	&	family	&	coordinate	\\
metal	&	mu	&	address	&	integral	&	hydride	&	egg	&	ribosomal	&	kidney	&	to	&	experiment	&	anti	&	box	&	mouse	\\
oxidoreductase	&	transit	&	storage	&	relation	&	extracellular	&	seed	&	terminus	&	corona	&	give	&	matter	&	approve	&	prism	&	silicon	\\
    \bottomrule
    \end{tabular}
\end{table}
\end{landscape}

\begin{table}[H]
    \singlespacing
    \begin{center}
     \caption{Top 30 Loadings of General Factors for Full Data}
     \label{tab:8}
    \footnotesize
    \begin{tabular}{ccccccccccccc}
    \toprule
$F_1$	&	$F_2$	&	$F_3$	&	$F_4$	&	$F_5$	&	$F_6$	&	$F_7$	&	$F_8$	&	$F_9$	&	$F_{10}$	&	$F_{11}$	&	$F_{12}$	&	$F_{13}$	\\
     \hline
0.750	&	0.744	&	0.748	&	0.771	&	0.671	&	0.700	&	0.629	&	0.655	&	-0.554	&	0.581	&	0.727	&	-0.570	&	-0.427	\\
0.742	&	0.740	&	0.729	&	0.729	&	0.660	&	0.632	&	0.617	&	0.640	&	-0.516	&	0.567	&	0.703	&	-0.534	&	-0.419	\\
0.742	&	0.732	&	0.720	&	0.689	&	0.653	&	0.630	&	0.612	&	0.625	&	-0.490	&	0.534	&	0.680	&	-0.513	&	-0.416	\\
0.739	&	0.718	&	0.717	&	0.668	&	0.641	&	0.600	&	0.606	&	0.602	&	-0.489	&	0.520	&	0.673	&	-0.487	&	0.398	\\
0.739	&	0.717	&	0.715	&	0.646	&	0.636	&	0.599	&	0.592	&	0.585	&	-0.483	&	0.515	&	0.654	&	-0.482	&	-0.391	\\
0.733	&	0.712	&	0.711	&	0.645	&	0.630	&	0.592	&	0.586	&	0.584	&	0.456	&	0.510	&	0.641	&	-0.477	&	-0.391	\\
0.728	&	0.704	&	0.706	&	0.640	&	0.624	&	0.588	&	0.582	&	0.577	&	-0.453	&	0.507	&	0.639	&	-0.472	&	-0.386	\\
0.725	&	0.699	&	0.702	&	0.628	&	0.610	&	0.587	&	0.580	&	0.568	&	0.429	&	0.505	&	0.638	&	-0.471	&	-0.382	\\
0.724	&	0.699	&	0.700	&	0.624	&	0.598	&	0.582	&	0.575	&	0.560	&	-0.427	&	0.502	&	0.607	&	-0.459	&	0.370	\\
0.714	&	0.697	&	0.699	&	0.620	&	0.596	&	0.581	&	0.567	&	0.553	&	-0.417	&	0.491	&	0.599	&	-0.438	&	0.368	\\
0.710	&	0.692	&	0.686	&	0.609	&	0.594	&	0.579	&	0.564	&	0.550	&	-0.417	&	0.482	&	0.589	&	-0.408	&	0.367	\\
0.710	&	0.684	&	0.683	&	0.607	&	0.586	&	0.577	&	0.561	&	0.543	&	-0.412	&	0.481	&	0.586	&	-0.401	&	-0.366	\\
0.710	&	0.681	&	0.678	&	0.597	&	0.581	&	0.577	&	0.561	&	0.540	&	-0.408	&	0.481	&	0.579	&	-0.397	&	0.363	\\
0.709	&	0.676	&	0.674	&	0.594	&	0.578	&	0.577	&	0.558	&	0.529	&	-0.406	&	0.480	&	0.559	&	-0.395	&	-0.362	\\
0.706	&	0.668	&	0.672	&	0.569	&	0.578	&	0.566	&	0.555	&	0.526	&	0.406	&	0.470	&	0.553	&	-0.395	&	-0.358	\\
0.705	&	0.665	&	0.665	&	0.567	&	0.576	&	0.564	&	0.551	&	0.523	&	0.404	&	0.467	&	0.550	&	-0.390	&	-0.349	\\
0.703	&	0.662	&	0.654	&	0.565	&	0.576	&	0.557	&	0.545	&	0.507	&	-0.404	&	0.461	&	0.538	&	-0.388	&	-0.348	\\
0.701	&	0.660	&	0.650	&	0.564	&	0.576	&	0.557	&	0.539	&	0.507	&	-0.402	&	0.458	&	0.533	&	-0.385	&	-0.347	\\
0.699	&	0.658	&	0.649	&	0.560	&	0.575	&	0.554	&	0.538	&	0.502	&	-0.401	&	0.455	&	0.528	&	-0.385	&	0.345	\\
0.697	&	0.655	&	0.644	&	0.559	&	0.566	&	0.545	&	0.529	&	0.501	&	-0.394	&	0.451	&	0.524	&	-0.385	&	0.343	\\
0.694	&	0.647	&	0.630	&	0.556	&	0.565	&	0.541	&	0.526	&	0.499	&	-0.393	&	0.444	&	0.520	&	-0.380	&	-0.343	\\
0.693	&	0.643	&	0.626	&	0.554	&	0.558	&	0.541	&	0.526	&	0.497	&	0.392	&	0.443	&	0.517	&	-0.378	&	-0.335	\\
0.691	&	0.641	&	0.624	&	0.553	&	0.555	&	0.540	&	0.521	&	0.496	&	-0.390	&	0.441	&	0.502	&	-0.371	&	-0.331	\\
0.690	&	0.634	&	0.614	&	0.548	&	0.551	&	0.532	&	0.512	&	0.492	&	-0.388	&	0.430	&	0.498	&	-0.369	&	0.330	\\
0.686	&	0.631	&	0.611	&	0.546	&	0.547	&	0.526	&	0.510	&	0.483	&	0.385	&	0.429	&	0.495	&	-0.365	&	-0.330	\\
0.685	&	0.631	&	0.611	&	0.535	&	0.546	&	0.518	&	0.509	&	0.482	&	-0.382	&	0.426	&	0.483	&	-0.363	&	0.329	\\
0.683	&	0.630	&	0.606	&	0.534	&	0.544	&	0.514	&	0.501	&	0.480	&	0.381	&	0.426	&	0.480	&	0.362	&	0.328	\\
0.681	&	0.625	&	0.605	&	0.532	&	0.543	&	0.512	&	0.488	&	0.474	&	-0.380	&	0.426	&	0.478	&	-0.359	&	0.328	\\
0.673	&	0.623	&	0.605	&	0.532	&	0.538	&	0.510	&	0.486	&	0.473	&	0.378	&	0.422	&	0.478	&	-0.349	&	-0.326	\\
0.667	&	0.617	&	0.603	&	0.531	&	0.536	&	0.505	&	0.485	&	0.471	&	0.377	&	0.421	&	0.469	&	0.348	&	0.324	\\
    \bottomrule
    \end{tabular}
    \end{center}
\end{table}

\begin{table}[H]
    \singlespacing
     \caption{Labels of General Factors}
     \label{tab:6}    
     \centering
    \setlength{\tabcolsep}{2pt} 
    \footnotesize
    \begin{tabular}{lll}
    \toprule
&	Full Data	&	Filtered	Data\\
     \hline
$F_1$	&	Chemical Elements and Compounds	&	Materials and Chemical Processes	\\
$F_2$	&	Astronomy and Celestial Objects	&	Computer Systems and Networking	\\
$F_3$	&	Computer Programming and Technology	&	Mathematical Concepts and Theories	\\
$F_4$	&	Mathematics and Mathematical Concepts	&	Biological Entities and Processes	\\
$F_5$	&	Biology and Biochemistry	&	Molecular Biology and Biochemistry	\\
$F_6$	&	Ecology and Organisms	& Astronomical Objects and Chemical Elements	\\
$F_7$	&	Genetics and Molecular Biology	&	Data Analysis and Signal Processing	\\
$F_8$	&	Anatomy and Physiology	&	Organisms and Ecology	\\
$F_9$	&	(Inverse) Biochemistry and Molecular Biology	&	Pharmacology and Drug Development	\\
$F_{10}$	&	Physics and Energy	&	Anatomy and Physiology	\\
$F_{11}$	&	Pharmacology and Drug Therapy	&	Classification and Relationships in Various Fields	\\
$F_{12}$	&	(Inverse) Enzyme Function and Metabolic Pathways	&	Biochemical Reactions and Molecular Interactions	\\
$F_{13}$	&	Medical Conditions and Molecular Processes	&	Transportation and Communication Systems	\\
    \bottomrule
    \end{tabular}
    {\\ \textit{Note}. Labels are suggested by ChatGPT based on top 30 keywords and corresponding loadings.}
\end{table}

\subsubsection{Classical Item Analysis}

With contextual scores, one can adopt the top words of different dimensions (i.e., factors) to construct scales simply based on classical psychometrics. Such scales, together with the LLM, can be used to "measure" the amount of knowledge any document contains. We employ top words for the first three factors as three sets of word items for illustration. As shown in Table \ref{tab:item_total}, the within-dimension and between-dimension item-total correlations are rather high and low, respectively, for most items. The coefficients are evidence of good convergent and discriminant power under the classical multi-trait multi-method approach, a indicator of valid measurement.

\begin{figure}
    \begin{center}
    \caption{Scatter and Density Plots of the Top Three Words for Item Set 1.}
    \label{fig:scatter3}    
    \includegraphics[width=.8\linewidth]{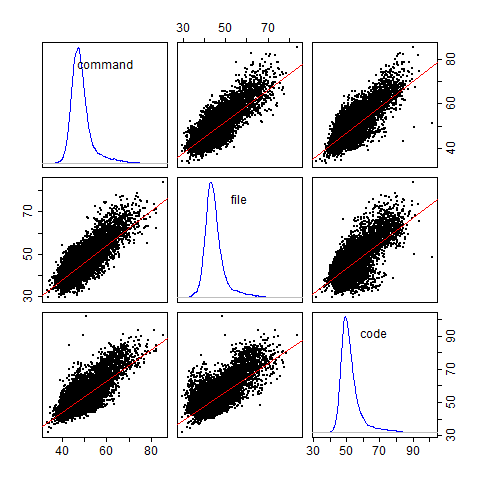}
    \end{center} 
\end{figure}

\begin{figure}
 \centering
    \caption{Scatter and Density Plots of Top Word for Three Item Sets.}
    \label{fig:scatter4}    
    \includegraphics[width=.8\linewidth]{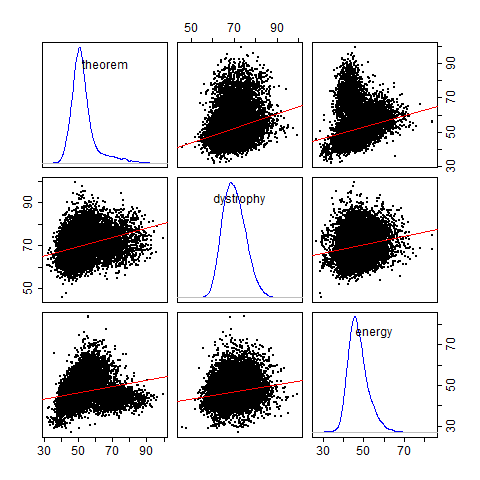}
    {\\ \textit{Note}. theorem - Set 1; dystrophy - Set 2; energy - Set 3.}
\end{figure}

The contextual scores' scatter and density plots for the top three words in the first item set can be found in Figure \ref{fig:scatter3} (Appendix F), which exhibits high correlations. Similar plots of three top words from the second and third sets can be found in Figure \ref{fig:scatter4} (Appendix F), which demonstrates weak correlations. Both figures give patterns consistent with the analytic results here.

\begin{table}[H]
    \singlespacing
\caption{Item Total Correlations} 
\label{tab:item_total}
 \centering
 \setlength{\tabcolsep}{2pt} 
 \footnotesize
\begin{tabular}{lcclcclcc}
    \toprule
Set 1 & Within & Between & Set 2 & Within & Between & Set 3 & Within & Between \\ 
\hline
sulfur & 0.93 & -0.18 & main & 0.90 & 0.09 & byte & 0.93 & -0.05 \\ 
zinc & 0.85 & -0.18 & constellation & 0.93 & -0.13 & software & 0.91 & -0.01 \\ 
  chloride & 0.90 & -0.33 & major & 0.84 & -0.07 & command & 0.83 & -0.05 \\ 
  sugar & 0.88 & -0.28 & star & 0.96 & -0.09 & file & 0.9 & -0.07 \\ 
  nitrogen & 0.90 & -0.23 & gamma & 0.80 & -0.04 & instruction & 0.89 & 0.01 \\ 
  amino & 0.86 & -0.11 & au & 0.89 & 0.09 & code & 0.80 & 0.10 \\ 
  vinyl & 0.87 & -0.19 & dwarf & 0.90 & -0.04 & program & 0.90 & 0.02 \\ 
  nickel & 0.81 & 0.12 & giant & 0.91 & -0.09 & server & 0.91 & -0.01 \\ 
  glucose & 0.82 & -0.28 & companion & 0.95 & 0.04 & java & 0.88 & 0.00 \\ 
  benzoate & 0.87 & -0.33 & kappa & 0.79 & 0.04 & processor & 0.89 & 0.10 \\ 
  chemistry & 0.91 & -0.15 & asteroid & 0.93 & -0.03 & library & 0.87 & 0.09 \\ 
  magnesium & 0.83 & -0.04 & catalogue & 0.85 & -0.01 & package & 0.87 & 0.12 \\ 
  amine & 0.88 & -0.34 & ngc & 0.92 & 0.03 & chip & 0.80 & -0.13 \\ 
  alcohol & 0.89 & -0.16 & spectral & 0.92 & -0.06 & programming & 0.83 & -0.28 \\ 
  bond & 0.81 & -0.16 & cluster & 0.77 & 0.12 & store & 0.90 & 0.06 \\ 
  carbon & 0.90 & -0.10 & milky & 0.87 & 0.03 & security & 0.86 & -0.04 \\ 
  ammonia & 0.83 & -0.16 & saros & 0.84 & -0.23 & computer & 0.9 & 0.13 \\ 
  aminoacyl & 0.81 & -0.35 & spectrum & 0.88 & -0.08 & linux & 0.82 & 0.25 \\ 
  benzene & 0.85 & -0.26 & jupiter & 0.81 & 0.18 & bit & 0.87 & 0.04 \\ 
  arsenic & 0.87 & -0.18 & hd & 0.88 & -0.01 & http & 0.84 & 0.00 \\ 
  phosphorus & 0.81 & -0.11 & planet & 0.89 & 0.10 & service & 0.79 & 0.00 \\ 
  oxide & 0.80 & 0.15 & component & 0.78 & 0.15 & format & 0.73 & 0.10 \\ 
  carbonyl & 0.86 & -0.30 & minor & 0.76 & 0.12 & cpu & 0.82 & 0.24 \\ 
  oxidation & 0.88 & -0.07 & super & 0.88 & 0.11 & machine & 0.87 & -0.07 \\ 
  aluminium & 0.78 & 0.22 & survey & 0.87 & -0.17 & key & 0.79 & 0.18 \\ 
  acid & 0.87 & -0.30 & beta & 0.57 & 0.16 & database & 0.79 & 0.08 \\ 
  oxygen & 0.88 & -0.08 & massive & 0.90 & 0.01 & memory & 0.79 & -0.14 \\ 
  nitrate & 0.86 & -0.19 & magnitude & 0.86 & -0.17 & unix & 0.86 & 0.17 \\ 
  metal & 0.76 & 0.18 & mu & 0.86 & 0.06 & address & 0.84 & -0.06 \\ 
  oxidoreductase & 0.87 & -0.35 & transit & 0.75 & 0.07 & storage & 0.85 & 0.20 \\ 
    \bottomrule
  \end{tabular}
{\\ \textit{Note}. Within: item-total correlations within the same dimension; Between: item-total correlations between different dimensions.}
\end{table}

\section{Discussion}

A key to LLMs' success is their ability to produce contextual embeddings, which encapsulate semantic relationships and contextual information within text. This research demonstrates that we can combine contextual embeddings with psychometric modeling to analyze textual data. By treating documents as individuals and words as items, this approach provides a natural psychometric interpretation with contextual scores serving as item responses. A crucial aspect of the approach is that contextual scores built upon contextual embeddings can be reliably obtained using different rephrasing forms. These scores also exhibit characteristics important for psychometric analysis: 1) Semantic relatedness: The score indicates the extent to which a word's meaning is related to the document's content. Words that are semantically relevant to the document will have higher contextual scores; 2) Normality and Linearity: The dot product generates a continuous score, enabling comparisons across different words and documents. Score distributions are approximately normal for most words, and the majority of word pairs exhibit linear relationships. This suggests that the scores are suitable for covariance structure analysis; 3) Stable distributions: The score distributions of the words are stable and unaffected by specific documents, which indicates that the words' characteristics, linked to the distributions, are independent of the documents. This aligns with the assumption of independence between items and individuals in psychometric modeling; 4) Latent patterns: Variations in contextual scores reflect latent patterns or hidden structure that can elucidate the relationships between words and documents, as well as the similarities or differences among documents within a corpus.

By modeling a corpus of documents using contextual scores and FA, the documents can be assigned with factor scores representing the latent constructs or knowledge. Although the numbers of first-order factors differ across the full and filtered data, second-order factors can be reliably extracted in the same number. The Bifactor structure with the SL transformation is preferred over the HO structure, since the general factors can be separated from a large number of minor factors which are usually of less interest. These factors can be directly defined and labeled substantively with significant keywords. The contextual scores can also be used for classical item analysis, and most keyword items exhibit good convergent and discriminant power.

The proposed method provides opportunities to facilitate psychometric analysis under the TDA context, such as item analysis, reliability, and validity assessments. Moreover, the scope of TDA can be further extended. For instance, one can analyze the factor correlations, score profiles of specific documents, score distributions of individual words, or correlations between words in a corpus. The approach can also be further expanded by including structural variables or predictors as covariates. This integration of additional variables would allow for a more comprehensive understanding of the relationships between the latent variables and external influences, thus enhancing the applicability and utility of the approach in various contexts.

However, the existing method is primarily exploratory due to the absence of hypothesis or theoretical support. Parameter and factor score estimates are deemed unreliable or inaccurate in exploratory FA, and exploratory Bifactor analysis is even less studied for these issues. On the other hand, there will be a large amount of model uncertainty in a complex textual environment even as knowledge and understanding accumulate over time, making a strictly confirmatory approach challenging. A more practical approach might be to incorporate the partially confirmatory factor analysis \parencite[e.g.,][]{chen2021partially,chen2022generalized}, where only a subset of loading patterns is specified based on prior knowledge, into the Bifactor model for better estimation of the model parameters and factor scores for the general factors. Specifically, one can choose some of the general factors identified, with the support from content experts, to facilitate a partially confirmatory approach. This approach can be further extended to incorporate various structural designs to better explain the factors \parencite{chen2024research}.

There are also other issues to address in future research. The contextual scores of many word pairs are highly correlated, which can be an important cause for the large number of minor factors together with some uncertainty in the general factors. One reason is that the BERT model and/or dot product might not be sufficiently sensitive to distinguish most words across different documents. Thus, the challenge is to find LLMs and/or other computation methods (e.g., cosine similarity) that can make the scores less correlated while keeping the good characteristics of the contextual scores.

Given these issues are solved, the proposed method can unveil a vast and unexplored landscape to analyze textual data with both LLMs and psychometric modeling, which will benefit researchers and practitioners in fields rich in textual data, such as education, psychology, and law. For instance, psychological counselors or lawyers can convert their textual data into contextual scores to uncover latent themes or to make statistical inferences through correlation or regression analysis. This approach can lead to substantial educational implications as textual data are ubiquitous and vital in education. When educational texts are measurable with factor score as knowledge, it will enable data-driven and knowledge-based curriculum design, assessment development, teaching preparation, and personalized learning. Moreover, researchers can align textual data with behavioral responses from different types of respondents through the same psychometric framework, which is still uncharted territory.

\printbibliography

\section{Appendix}
\subsection{A. First-order Scree Plot for Filtered Data}
\begin{figure}[H]
    \begin{center}
    \caption{Scree Plot of First-order FA for Full Data.}
    \label{fig:appa}    
    \includegraphics[width=.8\linewidth]{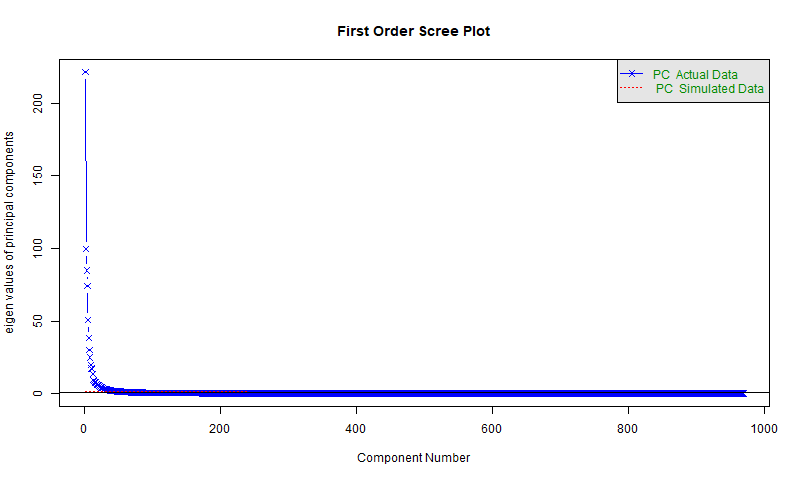}
    \end{center}
\end{figure}

\subsection{B. Second-order Scree Plot for Filtered Data}
\begin{figure}[H]
    \begin{center}
    \caption{Scree Plot of Second-Order FA for Full data.}
    \label{fig:appb}    
    \includegraphics[width=.8\linewidth]{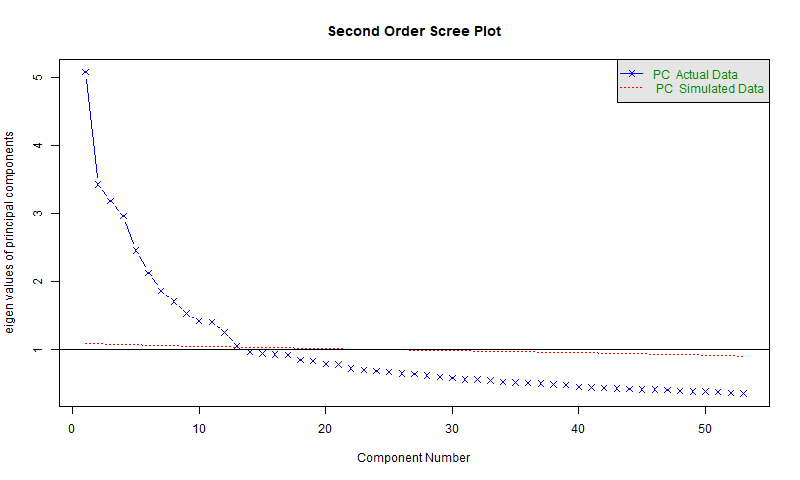}
    \end{center}
\end{figure}

\subsection{C. Second-order Factor Correlations for Filtered Data }
\begin{table}[H]
    \singlespacing
    \begin{center}
     \caption{Second-order Factor Correlations for Filtered Data}
     \label{tab:appc}
    \setlength{\tabcolsep}{3pt} 
    \small
    \begin{tabular}{lcccccccccccc}
    \toprule
 &	$Z_1$	&	$Z_2$	&	$Z_3$	&	$Z_4$	&	$Z_5$	&	$Z_6$	&	$Z_7$	&	$Z_8$	&	$Z_9$	&	$Z_{10}$	&	$Z_{11}$	&	$Z_{12}$	\\
     \hline
$Z_2$	&	0.110	&		&		&		&		&		&		&		&		&		&		&		\\
$Z_3$	&	0.144	&	0.179	&		&		&		&		&		&		&		&		&		&		\\
$Z_4$	&	0.086	&	0.072	&	-0.007	&		&		&		&		&		&		&		&		&		\\
$Z_5$	&	-0.052	&	-0.052	&	-0.070	&	0.034	&		&		&		&		&		&		&		&		\\
$Z_6$	&	0.155	&	0.038	&	0.075	&	0.045	&	0.055	&		&		&		&		&		&		&		\\
$Z_7$	&	-0.049	&	0.220	&	0.075	&	0.075	&	-0.095	&	0.085	&		&		&		&		&		&		\\
$Z_8$	&	-0.005	&	0.129	&	0.011	&	0.174	&	0.009	&	-0.015	&	0.138	&		&		&		&		&		\\
$Z_9$	&	0.097	&	0.008	&	-0.062	&	0.079	&	0.073	&	0.052	&	0.000	&	-0.037	&		&		&		&		\\
$Z_{10}$	&	0.117	&	0.112	&	0.085	&	0.135	&	-0.020	&	0.049	&	0.104	&	0.223	&	0.024	&		&		&		\\
$Z_{11}$	&	0.013	&	0.035	&	0.059	&	-0.030	&	-0.035	&	0.218	&	0.149	&	-0.023	&	-0.279	&	-0.004	&		&		\\
$Z_{12}$	&	0.070	&	0.018	&	0.065	&	-0.012	&	0.223	&	0.096	&	-0.062	&	-0.080	&	0.046	&	-0.066	&	0.067	&		\\
$Z_{13}$	&	-0.031	&	0.124	&	-0.020	&	0.035	&	0.104	&	0.097	&	0.189	&	0.053	&	-0.046	&	0.036	&	-0.027	&	-0.011	\\
    \bottomrule
    \end{tabular}
    \end{center}
\end{table}

\subsection{D. Top 30 Keywords of General Factors for Filtered Data }
\begin{landscape}
\begin{table}[H]
    \begin{center}
     \caption{Top 30 Words of General Factors for Filtered Data}
     \label{tab:appd}
    \renewcommand{\arraystretch}{1}
    \setlength{\tabcolsep}{0pt} 
    \footnotesize
    \begin{tabular}{ccccccccccccc}
    \toprule
$F_1$	&	$F_2$	&	$F_3$	&	$F_4$	&	$F_5$	&	$F_6$	&	$F_7$	&	$F_8$	&	$F_9$	&	$F_{10}$	&	$F_{11}$	&	$F_{12}$	&	$F_{13}$	\\
     \hline
explosive	&	code	&	theorem	&	streptomyces	&	subunit	&	gamma	&	event	&	colony	&	antagonist	&	nerve	&	diabetes	&	transfer	&	transport	\\
solvent	&	file	&	property	&	gastrointestinal	&	polymerase	&	zeta	&	variation	&	grow	&	selective	&	vascular	&	separation	&	acceptor	&	transporter	\\
electron	&	command	&	matrix	&	lymphoma	&	protein	&	alpha	&	cause	&	clade	&	drug	&	anterior	&	discover	&	local	&	pump	\\
metal	&	format	&	manifold	&	cholesterol	&	strand	&	delta	&	frequency	&	yeast	&	fda	&	ligament	&	binary	&	transferase	&	transporting	\\
uranium	&	service	&	finite	&	dystrophy	&	membered	&	omega	&	factor	&	organism	&	agent	&	neck	&	type	&	reaction	&	vehicle	\\
medium	&	http	&	continuous	&	vaginal	&	polymer	&	19	&	predict	&	gram	&	safety	&	lateral	&	genus	&	natural	&	signal	\\
glass	&	bit	&	differential	&	bacillus	&	molecule	&	component	&	error	&	mushroom	&	effect	&	cervical	&	open	&	de	&	force	\\
alloy	&	chip	&	axiom	&	cannabinoid	&	stranded	&	cluster	&	moment	&	coli	&	treatment	&	muscle	&	category	&	base	&	energy	\\
fuel	&	key	&	additive	&	pyruvate	&	transaminase	&	2021	&	detection	&	insect	&	trial	&	corona	&	pair	&	exchange	&	dynamic	\\
hydrate	&	address	&	solution	&	adrenergic	&	bond	&	star	&	delay	&	food	&	antibody	&	facial	&	blood	&	donor	&	movement	\\
hydra	&	unicode	&	coefficient	&	orthologs	&	amp	&	mg	&	loss	&	sea	&	brand	&	skin	&	position	&	to	&	carrier	\\
fe	&	message	&	graph	&	keratin	&	ligand	&	c1	&	transition	&	stem	&	exposure	&	bone	&	feature	&	name	&	receiver	\\
barium	&	port	&	case	&	neutrophil	&	terminus	&	na	&	occur	&	animal	&	psychedelic	&	tendon	&	cardiac	&	function	&	sensor	\\
isotope	&	ip	&	gradient	&	dehydrogenase	&	molecular	&	rho	&	marker	&	spore	&	analgesic	&	sensory	&	group	&	box	&	drive	\\
yield	&	user	&	theory	&	epinephelus	&	linker	&	cd	&	positive	&	fossil	&	action	&	medial	&	human	&	formula	&	release	\\
electrode	&	database	&	generate	&	immunoglobulin	&	pathway	&	10	&	indicator	&	fly	&	approve	&	motor	&	minor	&	cyclic	&	load	\\
liquid	&	access	&	linear	&	eridanus	&	binding	&	11	&	pattern	&	fish	&	substance	&	gland	&	compact	&	partial	&	transit	\\
map	&	memory	&	symmetric	&	mycobacterium	&	membrane	&	minor	&	measurement	&	male	&	use	&	central	&	pump	&	o2	&	plasma	\\
gas	&	country	&	kernel	&	carbohydrate	&	gtpase	&	1	&	negative	&	virus	&	antiviral	&	blood	&	dual	&	free	&	activity	\\
halide	&	string	&	set	&	pancreatic	&	homology	&	central	&	particle	&	population	&	resistance	&	prostate	&	semiregular	&	term	&	supply	\\
dye	&	run	&	symmetry	&	nonsteroidal	&	zinc	&	transit	&	stress	&	bacteria	&	analogue	&	nasal	&	like	&	short	&	channel	\\
green	&	language	&	calculus	&	neuron	&	repeat	&	carrier	&	frame	&	strain	&	attack	&	pain	&	object	&	intermediate	&	communication	\\
copper	&	client	&	knot	&	plasmodium	&	dna	&	12	&	shock	&	chromosome	&	risk	&	nervous	&	insulin	&	position	&	pressure	\\
oil	&	write	&	total	&	lysine	&	histone	&	sigma	&	shift	&	soil	&	inflammatory	&	fracture	&	hiv	&	end	&	track	\\
fluoride	&	support	&	transformation	&	hydride	&	degradation	&	group	&	experiment	&	leaf	&	extract	&	tube	&	cluster	&	residue	&	interaction	\\
domain	&	window	&	relation	&	hemoglobin	&	c1	&	binary	&	path	&	seed	&	fentanyl	&	plate	&	structure	&	2c	&	magnetic	\\
mercury	&	block	&	johnson	&	melanoma	&	mechanism	&	ch	&	change	&	egg	&	ester	&	heart	&	health	&	stranded	&	environment	\\
flavor	&	launch	&	prime	&	nematode	&	cytochrome	&	ring	&	resolution	&	genus	&	user	&	head	&	exoplanets	&	cis	&	activation	\\
co2	&	video	&	generator	&	hexagonal	&	ketone	&	primary	&	probability	&	jurassic	&	compound	&	junction	&	non	&	terminal	&	launch	\\
bromide	&	device	&	contraction	&	cartilage	&	phosphate	&	thyroid	&	focus	&	hepatitis	&	agonist	&	superior	&	phosphate	&	enzyme	&	event	\\
    \bottomrule
    \end{tabular}
    \end{center}
\end{table}
\end{landscape}

\subsection{E. Top 30 Loadings of General Factors Filtered Data }

\begin{table}[H]
    \singlespacing
    \begin{center}
     \caption{Top 30 Loadings of General Factors for Filtered Data}
     \label{tab:appe}
    \footnotesize
    \begin{tabular}{ccccccccccccc}
    \toprule
$F_1$	&	$F_2$	&	$F_3$	&	$F_4$	&	$F_5$	&	$F_6$	&	$F_7$	&	$F_8$	&	$F_9$	&	$F_{10}$	&	$F_{11}$	&	$F_{12}$	&	$F_{13}$	\\
     \hline
0.637	&	0.707	&	0.767	&	0.695	&	0.670	&	0.629	&	0.519	&	0.627	&	0.680	&	0.591	&	-0.463	&	0.646	&	0.583	\\
0.621	&	0.700	&	0.664	&	0.664	&	0.630	&	0.613	&	0.502	&	0.567	&	0.630	&	0.572	&	0.462	&	0.601	&	0.506	\\
0.610	&	0.695	&	0.660	&	0.648	&	0.605	&	0.604	&	0.489	&	0.559	&	0.618	&	0.566	&	0.439	&	0.517	&	0.441	\\
0.609	&	0.651	&	0.644	&	0.647	&	0.604	&	0.567	&	0.468	&	0.541	&	0.598	&	0.565	&	0.435	&	0.502	&	0.439	\\
0.598	&	0.639	&	0.633	&	0.638	&	0.594	&	0.532	&	0.464	&	0.533	&	0.573	&	0.547	&	0.431	&	0.490	&	0.430	\\
0.598	&	0.637	&	0.631	&	0.635	&	0.578	&	0.501	&	0.460	&	0.525	&	0.566	&	0.532	&	0.429	&	0.487	&	0.426	\\
0.590	&	0.637	&	0.595	&	0.634	&	0.574	&	0.478	&	0.447	&	0.520	&	0.553	&	0.507	&	0.426	&	0.480	&	0.406	\\
0.586	&	0.633	&	0.580	&	0.631	&	0.568	&	0.461	&	0.436	&	0.509	&	0.549	&	0.506	&	0.423	&	0.477	&	0.402	\\
0.581	&	0.633	&	0.575	&	0.622	&	0.542	&	0.458	&	0.432	&	0.508	&	0.545	&	0.501	&	0.419	&	0.464	&	0.402	\\
0.581	&	0.631	&	0.568	&	0.621	&	0.538	&	0.457	&	0.429	&	0.504	&	0.533	&	0.490	&	-0.409	&	0.462	&	0.401	\\
0.576	&	0.602	&	0.564	&	0.615	&	0.519	&	0.452	&	0.424	&	0.502	&	0.526	&	0.479	&	0.406	&	0.445	&	0.401	\\
0.576	&	0.596	&	0.547	&	0.614	&	0.514	&	0.445	&	0.418	&	0.499	&	0.521	&	0.478	&	0.403	&	0.433	&	0.396	\\
0.575	&	0.596	&	0.544	&	0.604	&	0.512	&	0.438	&	0.418	&	0.496	&	0.501	&	0.476	&	-0.400	&	0.429	&	0.378	\\
0.574	&	0.595	&	0.544	&	0.602	&	0.507	&	0.436	&	0.406	&	0.493	&	0.492	&	0.471	&	0.395	&	0.424	&	0.374	\\
0.570	&	0.594	&	0.542	&	0.601	&	0.506	&	0.435	&	0.404	&	0.483	&	0.490	&	0.463	&	-0.395	&	0.423	&	0.370	\\
0.566	&	0.594	&	0.538	&	0.599	&	0.504	&	0.432	&	0.404	&	0.482	&	0.485	&	0.458	&	0.394	&	0.419	&	0.363	\\
0.556	&	0.578	&	0.534	&	0.595	&	0.494	&	0.429	&	0.399	&	0.459	&	0.481	&	0.456	&	0.392	&	0.418	&	0.358	\\
-0.555	&	0.573	&	0.531	&	0.593	&	0.493	&	0.429	&	0.399	&	0.455	&	0.467	&	0.450	&	-0.392	&	0.412	&	0.358	\\
0.549	&	0.561	&	0.531	&	0.592	&	0.492	&	0.429	&	0.398	&	0.454	&	0.445	&	0.450	&	0.390	&	0.410	&	0.355	\\
0.549	&	0.553	&	0.530	&	0.588	&	0.492	&	0.414	&	0.398	&	0.452	&	0.444	&	0.449	&	0.390	&	0.409	&	0.354	\\
0.548	&	0.545	&	0.529	&	0.584	&	0.484	&	0.413	&	0.397	&	0.450	&	0.443	&	0.443	&	0.388	&	0.399	&	0.341	\\
0.545	&	0.545	&	0.527	&	0.584	&	0.483	&	0.412	&	0.395	&	0.447	&	0.443	&	0.441	&	0.383	&	0.399	&	0.336	\\
0.544	&	0.543	&	0.525	&	0.582	&	0.481	&	0.407	&	0.391	&	0.446	&	0.443	&	0.440	&	-0.379	&	0.393	&	0.330	\\
0.535	&	0.540	&	0.523	&	0.577	&	0.476	&	0.404	&	0.389	&	0.445	&	0.434	&	0.437	&	-0.376	&	0.390	&	0.320	\\
0.531	&	0.535	&	0.523	&	0.573	&	0.471	&	0.403	&	0.388	&	0.445	&	0.434	&	0.434	&	0.371	&	0.389	&	0.318	\\
-0.529	&	0.534	&	0.515	&	0.571	&	0.464	&	0.402	&	0.381	&	0.439	&	0.432	&	0.432	&	0.367	&	0.384	&	0.316	\\
0.527	&	0.529	&	0.514	&	0.570	&	0.461	&	0.401	&	0.378	&	0.435	&	0.426	&	0.425	&	-0.366	&	0.383	&	0.315	\\
0.525	&	0.528	&	0.512	&	0.568	&	0.460	&	0.400	&	0.377	&	0.434	&	0.425	&	0.424	&	0.362	&	0.382	&	0.312	\\
0.523	&	0.525	&	0.511	&	0.564	&	0.459	&	0.396	&	0.376	&	0.422	&	0.418	&	0.421	&	0.360	&	0.380	&	0.312	\\
0.522	&	0.522	&	0.511	&	0.549	&	0.459	&	0.392	&	0.373	&	0.418	&	0.418	&	0.417	&	-0.359	&	0.379	&	0.311	\\
    \bottomrule
    \end{tabular}
    \end{center}
\end{table}

\subsection{F. Scatter and Density Plots for Three Item Sets}

\end{document}